\documentclass{article}
\usepackage{graphicx}
\usepackage{booktabs} 
\usepackage{scalerel}
\usepackage{float}

\usepackage{subcaption}
\usepackage{caption}
\usepackage[toc,page]{appendix}

\usepackage{url}            
\usepackage{amsfonts}       
\usepackage{nicefrac}       
\usepackage{microtype}      
\usepackage{cancel}

\usepackage[dvipsnames]{xcolor}
\usepackage{amsmath, latexsym}
\usepackage[most]{tcolorbox}
\usepackage{placeins}
\usepackage{enumitem}
\usepackage{bold-extra}
\usepackage[T1]{fontenc}
\usepackage{standalone}
\usepackage{tikz}
\usepackage{tikzscale}
\usepackage{pgfplots}
\usepackage{pgf}
\usetikzlibrary{calc}
\usetikzlibrary{positioning}
\usetikzlibrary{angles,quotes}
\usetikzlibrary{backgrounds}
\usetikzlibrary{external}
\usetikzlibrary{fit}
\usetikzlibrary{arrows}
\usetikzlibrary{arrows.meta}
\usetikzlibrary{shapes.symbols}
\usetikzlibrary{shadings}
\usetikzlibrary{shapes}
\usetikzlibrary{fadings}
\usetikzlibrary{bayesnet}
\usetikzlibrary{matrix}
\usetikzlibrary{snakes}
\usetikzlibrary{decorations.pathmorphing, patterns}
\pgfplotsset{compat=1.15}

\newcommand{\Rect}[5]{
    \draw[#1] (#2,#3) rectangle(#2+#4,#3-#5);
}

\definecolor{C1}{HTML}{1F77B4}
\definecolor{C2}{HTML}{FF7F0E}
\definecolor{C3}{HTML}{2CA02C}
\definecolor{C4}{HTML}{D62728}
\definecolor{C5}{HTML}{9467BD}
\colorlet{C1light}{C1!70!white}
\colorlet{C2light}{C2!70!white}
\colorlet{C3light}{C3!70!white}
\colorlet{C4light}{C4!70!white}
\colorlet{C5light}{C5!70!white}
\colorlet{C1vlight}{C1!20!white}
\colorlet{C2vlight}{C2!20!white}
\colorlet{C3vlight}{C3!20!white}
\colorlet{C4vlight}{C4!20!white}
\colorlet{C5vlight}{C5!20!white}
\colorlet{linkcolor}{violet}
\colorlet{citecolor}{RedOrange}  %
\colorlet{urlcolor}{Aquamarine}
\definecolor{ultrapink}{rgb}{1.0, 0.44, 1.0}
\RequirePackage{contour}
\RequirePackage{xspace}

\tikzset{
    /pgf/decoration/amplitude = 0.1em,
    /pgf/decoration/segment length = 0.5em}
\usepackage{cuted}
\usepackage{flushend}

\usepackage{hyperref}

\usepackage{nccmath}
\usepackage[accepted]{icml2021}

\usepackage{xargs}                      
\usepackage[colorinlistoftodos,prependcaption,textsize=tiny]{todonotes}
\newcommandx{\unsure}[2][1=]{\todo[linecolor=red,backgroundcolor=red!25,bordercolor=red,#1]{#2}}
\newcommandx{\change}[2][1=]{\todo[linecolor=blue,backgroundcolor=blue!25,bordercolor=blue,#1]{#2}}
\newcommandx{\info}[2][1=]{\todo[linecolor=OliveGreen,backgroundcolor=OliveGreen!25,bordercolor=OliveGreen,#1]{#2}}
\newcommandx{\improvement}[2][1=]{\todo[linecolor=Plum,backgroundcolor=Plum!25,bordercolor=Plum,#1]{#2}}
\newcommandx{\thiswillnotshow}[2][1=]{\todo[disable,#1]{#2}}

\usepackage{mdframed}
\newmdenv[
  topline=false,
  bottomline=false,
  rightline=false,
  skipabove=\topsep,
  ]{siderules}
  
\newcommand{\gparam}{\boldsymbol{\tau}}

\newcommand{\gspace}{\Omega_\tau}

\newcommand{\ngrad}{\mbox{$\hat{g}$}}
\newcommand{\vngrad}{\mbox{$\myvecsym{\ngrad}$}}

\newcommand{\vlat}{\vw}
\newcommand{\lat}{w}

\newcommand{\entropy}{\mathcal{H}}

\newcommand\cut[1]{}

\newcommand{\squishlist}{
   \begin{list}{$\bullet$}
    { \setlength{\itemsep}{0pt}      \setlength{\parsep}{3pt}
      \setlength{\topsep}{3pt}       \setlength{\partopsep}{0pt}
      \setlength{\leftmargin}{1.5em} \setlength{\labelwidth}{1em}
      \setlength{\labelsep}{0.5em} } }

\newcommand{\squishlisttwo}{
   \begin{list}{$\bullet$}
    { \setlength{\itemsep}{0pt}    \setlength{\parsep}{0pt}
      \setlength{\topsep}{0pt}     \setlength{\partopsep}{0pt}
      \setlength{\leftmargin}{2em} \setlength{\labelwidth}{1.5em}
      \setlength{\labelsep}{0.5em} } }

\newcommand{\squishend}{
    \end{list}  }

{}
{}
{}
{}
{}
{}
{}

\newcommand{\half}{\mbox{$\frac{1}{2}$}}

\newcommand{\real}{\mbox{$\mathbb{R}$}}

\newcommand{\rnd}[1]{\left(#1\right)}
\newcommand{\sqr}[1]{\left[#1\right]}

\newcommand{\myexpect}{\mathbb{E}}
\newcommand{\Unmyexpect}[1]{\mathbb{E}_{\scaleto{#1\mathstrut}{6pt}}}

\newcommand{\gauss}{\mbox{${\cal N}$}}

\newcommand{\myvec}[1]{\mbox{$\mathbf{#1}$}}
\newcommand{\myvecsym}[1]{\mbox{$\boldsymbol{#1}$}}

\newcommand{\vmu}{\mbox{$\myvecsym{\mu}$}}

\newcommand{\vSigma}{\mbox{$\myvecsym{\Sigma}$}}

\newcommand{\vg}{\mbox{$\myvec{g}$}}
\newcommand{\vh}{\mbox{$\myvec{h}$}}

\newcommand{\vw}{\mbox{$\myvec{w}$}}

\newcommand{\vy}{\mbox{$\myvec{y}$}}

\newcommand{\vB}{\mbox{$\myvec{B}$}}
\newcommand{\vC}{\mbox{$\myvec{C}$}}

\newcommand{\vF}{\mbox{$\myvec{F}$}}
\newcommand{\vG}{\mbox{$\myvec{G}$}}

\newcommand{\vI}{\mbox{$\myvec{I}$}}
\newcommand{\vJ}{\mbox{$\myvec{J}$}}
\newcommand{\vK}{\mbox{$\myvec{K}$}}

\newcommand{\vM}{\mbox{$\myvec{M}$}}

\newcommand{\vS}{\mbox{$\myvec{S}$}}

\newcommand{\vU}{\mbox{$\myvec{U}$}}

\newcommand{\vX}{\mbox{$\myvec{X}$}}

\newcommand{\be}{\begin{equation}}
\newcommand{\ee}{\end{equation}}
\newcommand{\bea}{\begin{eqnarray}}
\newcommand{\eea}{\end{eqnarray}}
\newcommand{\beaa}{\begin{eqnarray*}}
\newcommand{\eeaa}{\end{eqnarray*}}

\begin{document}

\newcommand{\ourtitle}{
Structured second-order methods via natural-gradient descent}

\icmltitlerunning{\ourtitle}

\twocolumn[

\icmltitle{\ourtitle}




\vspace{-0.31cm}
\begin{icmlauthorlist}
\icmlauthor{Wu Lin}{ubc}
\icmlauthor{Frank Nielsen}{sony}
\icmlauthor{Mohammad Emtiyaz Khan}{riken}
\icmlauthor{Mark Schmidt}{ubc,amii}
\end{icmlauthorlist}

\icmlaffiliation{ubc}{University of British Columbia.}
\icmlaffiliation{riken}{RIKEN Center for Advanced Intelligence Project.}
\icmlaffiliation{sony}{Sony Computer Science Laboratories Inc.}
\icmlaffiliation{amii}{CIFAR AI Chair, Alberta Machine Intelligence Institute}

\icmlcorrespondingauthor{Wu Lin}{yorker.lin@gmail.com
\vspace{-0.2cm}
}

\icmlkeywords{Natural Gradient Descent, Information Geometry, Variational Inference, Optimization, Search, Deep Learning}

\vskip 0.3in
]



\printAffiliationsAndNotice{}  

\begin{abstract}
We propose new structured second-order methods and structured adaptive-gradient methods obtained by performing natural-gradient descent on structured parameter spaces.
Natural-gradient descent is an attractive approach to design new algorithms in many settings such as gradient-free, adaptive-gradient, and second-order methods.
Our structured methods not only enjoy a structural invariance but also admit a simple expression.
Finally, we test the efficiency of our proposed methods on both deterministic non-convex problems and deep learning problems.
\vspace{-0.3cm}

\end{abstract}

\section{Introduction}

Newton's method is a powerful optimization method and is invariant to any invertible linear transformation.
Unfortunately, it is computationally intensive due to the inverse of the Hessian computation.
Moreover, it could perform poorly in non-convex settings since the Hessian matrix can be neither positive-definite nor invertible.
To address these issues, structural extensions  are proposed such as BFGS. However, these extensions often lose the linear invariance.
Moreover, existing extensions are often limited to one kind of structures and may not perform well in stochastic settings.  

In this paper, we introduce new efficient and structured 2nd-order updates that incorporate flexible structures and preserve a structural invariance. Unlike existing second-order methods, our methods can be readily  used in non-convex and stochastic settings.
Moreover,
structured and efficient adaptive-gradient methods are easily obtained for deep leaning  by using the Gauss-Newton approximation of the Hessian.
This work is an extension and application of the structured natural-gradient method~\citep{lin2021tractable}. 

Many machine learning applications  can be expressed as the following unconstrained optimization problem.

\vspace{-0.33cm}
\begin{align}
\min_{\mu \in \text{\real}^p} \ell(\vmu) \label{eq:opt}
\end{align}

\vspace{-0.33cm}
where 
function $\ell$ is a loss function.
Instead of directly solving \eqref{eq:opt}, we consider to solve the following optimization problem over a probabilistic distribution $q$.

\vspace{-0.32cm}
\begin{align}
   \min_{\gparam \in \gspace} \Unmyexpect{q(\text{\vlat}|\gparam)} \sqr{ \ell(\vlat) } - \gamma\entropy (q(\vlat|\gparam)) \label{eq:problem},
\end{align}

\vspace{-0.32cm}
where  $q(\vlat|\gparam)$ is a parametric distribution with parameters $\gparam$ in the parameter space $\gspace$, $\entropy(q(\vlat))$ is Shannon's entropy,  and $\gamma\ge 0$ is a constant.
Problem  \eqref{eq:problem} arises in many settings such as
gradient-free problems~\citep{baba1981convergence,beyer2001theory,spall2005introduction},
reinforcement learning~\citep{sutton1998introduction,williams1991function,teboulle1992,mnih2016asynchronous},
Bayesian inference~\citep{zellner1986bayesian},
and robust or global optimization~\citep{mobahi2015theoretical, leordeanu2008smoothing, hazan2016graduated}.

Natural-gradient descent (NGD) is an attractive method  to solve \eqref{eq:problem}. 
A standard NGD update with  step-size $\beta>0$ is

\vspace{-0.32cm}
\begin{equation}
\gparam_{t+1} \leftarrow \gparam_t  - \beta \vngrad_{\gparam_t}
\end{equation}

\vspace{-0.32cm}
where natural gradients $ \vngrad_{\gparam_t}$ are computed as below.

\vspace{-0.32cm}
\begin{equation*}
\vngrad_{\gparam_t}:= \vF_{\gparam}(\gparam_t)^{-1} \nabla_{\gparam}\sqr{ \myexpect_{q(\text{\vlat}|\gparam_t)} \sqr{ \ell(\vlat)} - \gamma\entropy (q(\vlat|\gparam_t))}
\end{equation*}

\vspace{-0.32cm}
and $\vF_{\gparam}(\gparam)\mspace{-6mu}:=\mathbb{E}_{q}[\nabla_\tau\log q(\vlat|\gparam)(\nabla_\tau^\top\log q(\vlat|\gparam))] $ is the Fisher information matrix (FIM).

\citet{khan2017variational,khan18a} show a connection between the standard NGD for \eqref{eq:problem} and Newton's method for \eqref{eq:opt}, when
 $q(\vlat|\gparam)$ is a Gaussian distribution $\gauss(\vlat|\vmu,\vS^{-1})$ with $\gparam=\{\vmu,\vS\}$, where $\vmu$ is the mean and $\vS$ is the precision.
 
\vspace{-0.32cm}
\begin{equation}
\begin{split}
\vmu_{t+1} & \leftarrow \vmu_t - \beta \vS_{t+1}^{-1} \myexpect_{q(\text{\vlat}|\gparam_t)}{ \sqr{ \nabla_\lat \ell( \vlat) }},  \\ 
\vS_{t+1} & \leftarrow (1-\beta\gamma)\vS_t + \beta  \Unmyexpect{q(\text{\vlat}|\gparam_t)}{ \sqr{ \nabla_\lat^2 \ell(\vlat) } }, \label{eq:newton_update}
\end{split}
\end{equation}

\vspace{-0.32cm}
The standard Newton's update for
problem \eqref{eq:opt} is recovered by approximating the expectations at the mean $\vmu$ and using  step-size $\beta=1$ when $\gamma =1$~\citep{emti2020bayesprinciple}.


Since the precision $\vS$ lies in a positive-definite matrix space, the update \eqref{eq:newton_update} may violate the constraint~\citep{khan18a}.
\citet{lin2020handling} propose an extension by introducing a correction term.
With this modification, we obtain a Newton-like update in \eqref{eq:gauss_rgd} for stochastic and non-convex problems.

\vspace{-0.32cm}
\begin{equation}
\begin{split}
  \mspace{-10.5mu} \vmu_{t+1} \mspace{-2.5mu} & \mspace{-2.5mu} \leftarrow \vmu_t - \beta \vS_{t}^{-1} \myexpect_{q_t}{\sqr{ \nabla_\lat \ell( \vlat) }},  \\ 
 \mspace{-10.5mu} \vS_{t+1} \mspace{-2.5mu}& \mspace{-2.5mu} \leftarrow (1-\beta\gamma) \vS_t + \beta \Unmyexpect{q_t}\mspace{-4mu}\sqr{\nabla_\lat^2 \ell( \vlat)} {+ \color{red} \frac{\beta^2}{2}  \vG_t \vS_{t}^{-1} \vG_t },
\end{split}
\mspace{-6.5mu}
\label{eq:gauss_rgd}
\end{equation}

\begin{figure*}[t]
\captionsetup[subfigure]{aboveskip=-1pt,belowskip=-1pt}
        \begin{subfigure}[b]{0.1\textwidth}
        \includegraphics[width=1\linewidth]{upper_triangular.tikz}
        \caption{}
	     \label{fig:group1}
	      \end{subfigure}
	      \hspace{0.16cm}
        \begin{subfigure}[b]{0.1\textwidth}
        \includegraphics[width=1\linewidth]{lower_triangular.tikz}
        \caption{}
	      \label{fig:group2}
	      \end{subfigure}
	      \hspace{0.16cm}
        \begin{subfigure}[b]{0.1\textwidth}
        \includegraphics[width=1\linewidth]{Heisenberg.tikz}
        \caption{}
	      \label{fig:group3}
	\end{subfigure}
	      \hspace{0.16cm}
        \begin{subfigure}[b]{0.1\textwidth}
        \includegraphics[width=1\linewidth]{conjugation.tikz}
              \caption{}
	\label{fig:group4}
	 \end{subfigure}
	     \hspace{0.16cm}
        \begin{subfigure}[b]{0.1\textwidth}
        \includegraphics[width=1\linewidth]{sparsecholesky.tikz}
              \caption{}
	\label{fig:group6}
	 \end{subfigure} 
	 	     \hspace{0.16cm}
        \begin{subfigure}[b]{0.1\textwidth}
        \includegraphics[width=1\linewidth]{TriToeplitz.tikz}
              \caption{}
	\label{fig:group7}
	 \end{subfigure}
	      \hspace{0.16cm}
        \begin{subfigure}[b]{0.28\textwidth}
	 \includegraphics[width=0.98\linewidth]{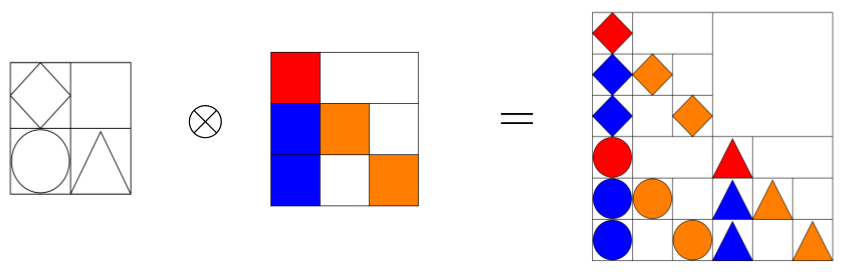} 
              \caption{}
	\label{fig:group5}
	 \end{subfigure}
	 \vspace{-0.2cm}
           \caption{ 
           Visualization of
         some useful  group structures.
         Figure~\ref{fig:group1} is a block upper-triangular group with $k=2$.
         Figure~\ref{fig:group2} is a block lower-triangular group with $k=2$.
         Figure~\ref{fig:group3} is a block Heisenberg group with $k_1=2$ and $k_2=4$, which is a hierarchical extension of Figure~\ref{fig:group2}.
         Figure~\ref{fig:group4} is a group conjugation of Figure~\ref{fig:group2} by a permutation matrix.
         Figure~\ref{fig:group6} is a sparse Cholesky group.
         Figure~\ref{fig:group7} is a triangular-Toeplitz group.
         Figure~\ref{fig:group5} is a Kronecker product group, which is a Kronecker product of two (block) lower-triangular groups.
         }
	\label{fig:groups}
\vspace{-0.4cm}
\end{figure*}

\vspace{-0.3cm}
where $\vG_t = \Unmyexpect{q_t}\sqr{\nabla_\lat^2 \ell( \vlat)} - \gamma \vS_t$
and $q_t(\vlat):=\gauss(\vlat|\vmu_t,\vS_t)$.
By setting  $0<\beta<1$ and $\gamma=1$, we obtain a second-order update  with a moving average on $\vS$, where the correction  colored in red is added  to handle the positive-definite constraint.
\citet{lin2020handling} study \eqref{eq:gauss_rgd} in doubly stochastic settings (e.g., stochastic variational inference using \eqref{eq:problem}), where noise comes from the mini-batch sampling and the Monte Carlo approximation to the expectations.

The structured natural-gradient method 
generalizes several NGD methods  including gradient-free methods~\citep{glasmachers2010exponential} and these second-order methods ~\citep{lin2020handling,khan18a}.
We could obtain  adaptive-gradient methods from a second-order update by using the Gauss-Newton approximation or randomized linear algebra to approximate the Hessian.
In this paper,
we show that new structured second-order methods for \eqref{eq:opt} can be obtained by performing NGD for \eqref{eq:problem} with structured Gaussians.
We further show that these structured methods preserve  a structural  invariance.
Finally, we test these structured methods on problems of numerical optimization and deep learning.

\label{sec:local_params}

\vspace{-0.2cm}
\section{Tractable NGD for structured Gaussians}
The structured natural-gradient method~\citep{lin2021tractable} is a systematic approach to incorporate flexible structured covariances with simple updates.
We will show that structured second-order methods for problem \eqref{eq:opt} with a structural invariance can be easily derived from these NGD updates.

We use
$\mathcal{S}_{++}^{p\times p}$, $\mathcal{S}^{p\times p}$, and $\mathrm{GL}^{p\times p}$ to denote
the set of symmetric positive definite matrices, symmetric matrices, and invertible matrices, respectively.
We define  map $\vh()$ on matrix $\vM$ as $\vh(\vM):=\vI+\vM+\half \vM^2$.
\vspace{-0.1cm}

\subsection{Full  covariance}
\label{sec:GaussSquareRoot}
We consider a $p$-dimensional Gaussian distribution $q(\vlat)=\gauss(\vlat|\vmu,\vS^{-1})$, where $\vS \in \mathcal{S}_{++}^{p\times p}$ is the precision matrix and $\vSigma=\vS^{-1}$ is the covariance matrix.
We consider the following covariance structure $\vS=\vB\vB^T$, where $\vB \in \mathrm{GL}^{p\times p}$ is an invertible matrix.
Using the structured  NGD method,  the update for \eqref{eq:problem} is expressed  as:
\vspace{-0.1cm}
\begin{equation}
    \begin{split}
    \vmu_{t+1} &\leftarrow \vmu_t - \beta \vS_{t}^{-1} \vg_{\mu_t} \\
    \vB_{t+1} &\leftarrow \vB_t \mathbf{h}\big(\beta \vB_t^{-1}\vg_{\Sigma_t} \vB_t^{-T} \big),
    \end{split}
     \label{eq:ngd_aux_param}
 \end{equation} 
where
$\vg_{\mu}$ and $\vg_{\Sigma}$ are vanilla gradients for $\vmu$ and $\vSigma$.  We can show in \eqref{eq:ngd_aux_param}, $\vB_{t+1} \in \mathrm{GL}^{p\times p}$  whenever $\vB_t \in \mathrm{GL}^{p\times p}$.
 Thus, $\vB_{t+1}$ is invertible if initial $\vB_0$ is invertible.
 
By {Stein's identity}~\citep{opper2009variational}, we have
\vspace{-0.1cm}
\begin{equation}
    \begin{split}
\vg_\mu = \Unmyexpect{q}{ \sqr{ \nabla_\lat \ell( \vlat) } }, \,\,\,\,\,  
\vg_{\Sigma}
 = \half \Unmyexpect{q}{ \sqr{ \nabla_\lat^2 \ell( \vlat) } } -\frac{\gamma}{2} \vS 
 \end{split}
 \label{eq:gauss_stein_2nd}
 \end{equation}
Using $\vS=\vB\vB^T$ and \eqref{eq:gauss_stein_2nd},
the update in \eqref{eq:ngd_aux_param} can be expressed as a second-order update as \eqref{eq:gauss_rgd} in terms of $\vmu$ and $\vS$.

\vspace{-0.35cm}
\begin{align*}
    \vmu_{t+1} &\leftarrow  \vmu_t - \beta \vS_{t}^{-1} \Unmyexpect{q_t}{ \sqr{ \nabla_\lat \ell( \vlat) } } \\
\vS_{t+1}  &\leftarrow  (1-\beta \gamma)\vS_t + \beta \Unmyexpect{q_t}\sqr{\nabla_\lat^2 \ell( \vlat)} {+ \color{red} \frac{\beta^2}{2}  \vG_t \vS_{t}^{-1} \vG_t } + O(\beta^3) \label{eq:gauss_newton_exp}
\end{align*}

\vspace{-0.35cm}
where $O(\beta^3)$ contains higher order terms.

Using \eqref{eq:gauss_stein_2nd} and
approximating the expectations in \eqref{eq:ngd_aux_param} at the mean $\vmu$, we obtain an update for
\eqref{eq:opt}, where $\vS=\vB\vB^T$.
\begin{tcolorbox}[enhanced,colback=white,%
    colframe=red!75!black, attach boxed title to top right={yshift=-\tcboxedtitleheight/2, xshift=-1.25cm}, title=A Newton-like update for \eqref{eq:opt}, coltitle=red!75!black, boxed title style={size=small,colback=white,opacityback=1, opacityframe=0}, size=title, enlarge top initially by=-\tcboxedtitleheight/2]
    \vspace{0.2cm}
\begin{equation}
    \begin{split}
 \vmu_{t+1} &\leftarrow \vmu_t - \beta \vS_{t}^{-1} \nabla_\mu \ell( \vmu_t) \\
    \vB_{t+1} &\leftarrow \vB_t \mathbf{h}\big(\frac{\beta}{2} \vB_t^{-1} ( \nabla_\mu^2 \ell( \vmu_t) -\gamma \vS_t ) \vB_t^{-T} \big)\\
    &= \vB_t \mathbf{h}\big(\frac{\beta}{2} ( \vB_t^{-1}  \nabla_\mu^2 \ell( \vmu_t)  \vB_t^{-T} -\gamma \vI) \big).
    \end{split}
 \label{eq:full_newton}
 \end{equation} 
\end{tcolorbox}

\vspace{-0.25cm}
Since  \eqref{eq:opt} only contains $\vmu$, we can view $\vB$ in \eqref{eq:full_newton}  as an axillary variable induced by the (hidden) geometry of Gaussian distribution.
Compared to the classical geometric interpretation for Newton's method,
this new view  allows us to handle the positive-definite constraint and exploit structures in $\vB$.

Moreover, update  \eqref{eq:full_newton} is invariant to any invertible linear transformation like Newton's method for \eqref{eq:opt}.
Consider a linear transformation $\vmu = \vK \vy$ of the variable $\vmu$ in the original  problem \eqref{eq:opt}, where $\vK \in \mathrm{GL}^{p \times p} $ is an known invertible  matrix. 
Let $f(\vy):=\ell(\vK \vy)$  be a new loss function.
The update for a new problem $\min_{y \in \text{\real}^p} f(\vy)$ is

\vspace{-0.35cm}
\begin{equation*}
    \begin{split}
    \vy_{t+1} &\leftarrow \vy_t - \beta \vC_{t}^{-T} \vC_t^{-1} \nabla_y f( \vy_t) \\
    \vC_{t+1} &\leftarrow
\vC_t \mathbf{h}\big(\frac{\beta}{2} ( \vC_t^{-1}  \nabla_y^2 f( \vy_t)  \vC_t^{-T} -\gamma \vI) \big).
    \end{split}
 \end{equation*} 
 
\vspace{-0.35cm}
By the construction, we have $\vmu_0 \mspace{-4mu}= \mspace{-4mu}\vK \vy_0$.
Since $\vK$ is known, we can initialize $\vC_0$ and $\vB_0$ so that $\vC_0 \mspace{-4mu}=\mspace{-4mu}\vK^T \vB_0 \mspace{-3mu}\in \mspace{-3mu}\mathrm{GL}^{p \times p}$.

By induction,  the following relationships hold. 
 $\vC_{t+1} = \vK^T \vB_{t+1}$ and $\vmu_{t+1}=\vK\vy_{t+1}$ for any $t\geq 0$.
 
Since  $\vmu_{t+1}=\vK\vy_{t+1}$, we have $f(\vy_{t+1})=\ell(\vK \vy_{t+1})=\ell(\vmu_{t+1})$. This  expression 
 shows that the  update on $f$ is also the same as that on $\ell$ at each iteration. Thus, this update is invariant to any invertible linear transformation $\vK \mspace{-3mu} \in  \mspace{-3mu}\mathrm{GL}^{p \times p}$.



\begin{figure*}[t]
\captionsetup[subfigure]{aboveskip=-1pt,belowskip=-1pt}
        \centering
\hspace*{-1.5cm}
        \begin{subfigure}[b]{0.25\textwidth}
	\includegraphics[width=\textwidth]{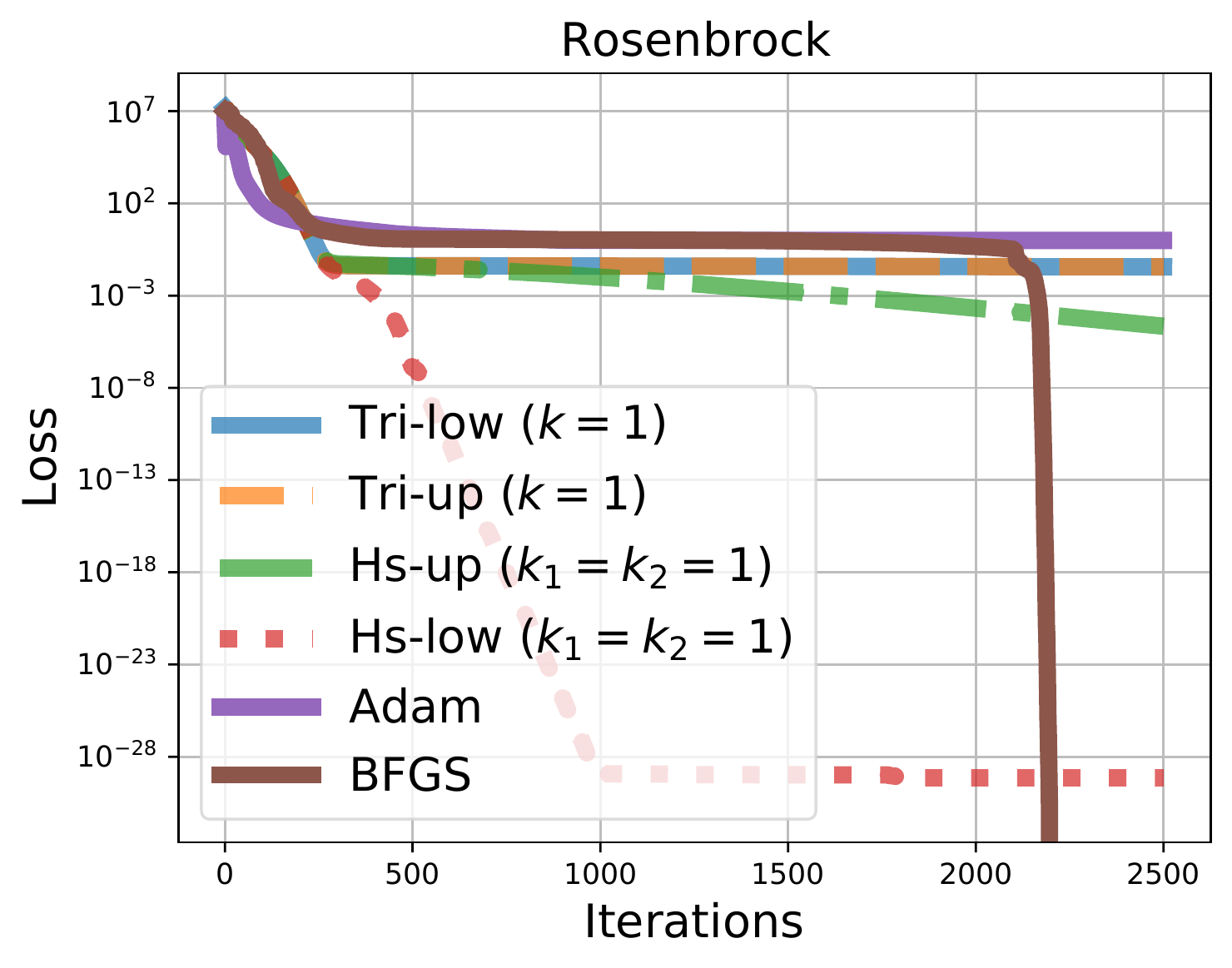}
                \caption{}
	      \label{fig:b}
        \end{subfigure}
	\hspace*{-0.2cm}
         \begin{subfigure}[b]{0.25\textwidth}
	\includegraphics[width=\textwidth]{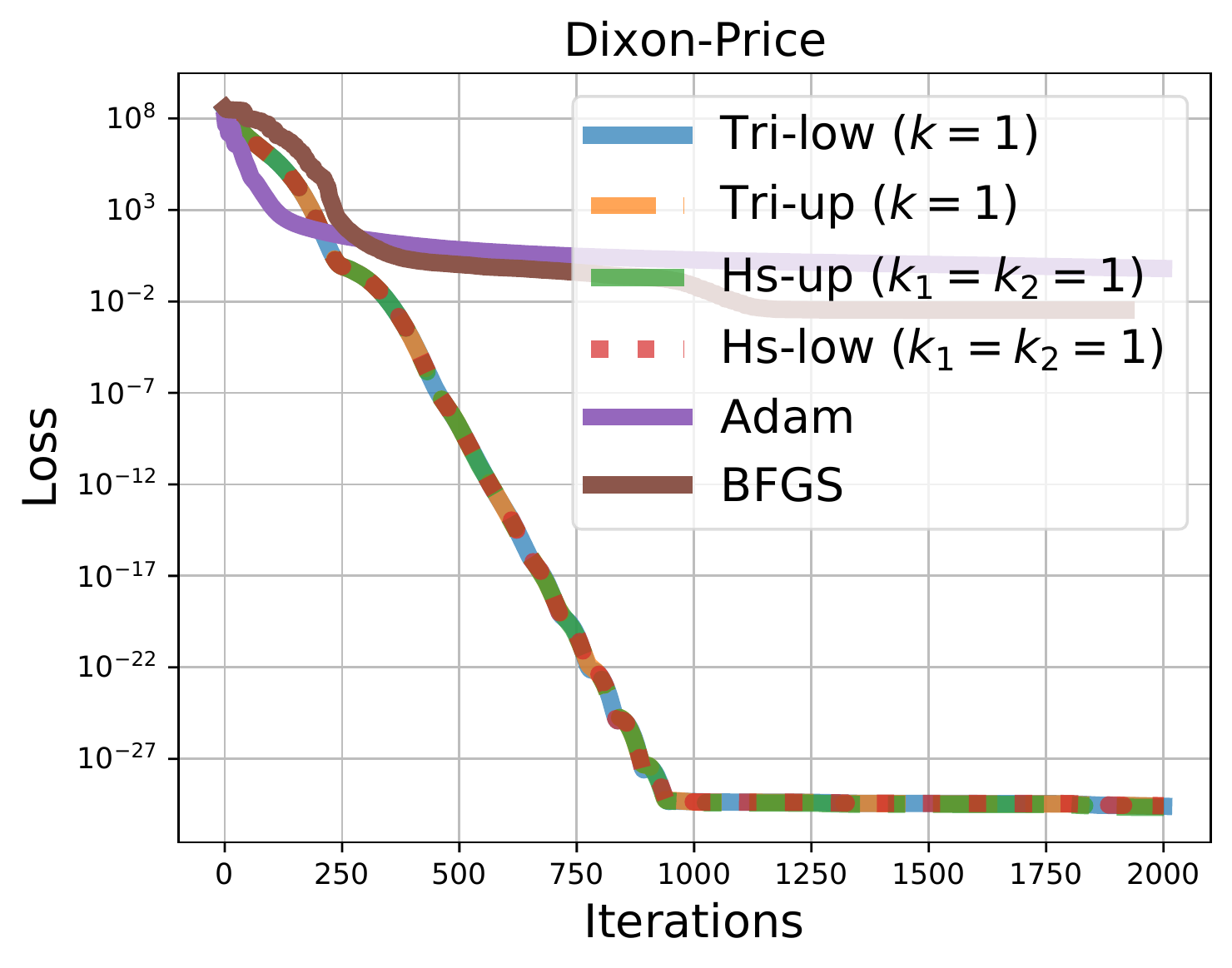}
                \caption{}
	      \label{fig:c}
        \end{subfigure}
	\hspace*{-0.2cm}
        \begin{subfigure}[b]{0.25\textwidth}
	\includegraphics[width=\textwidth]{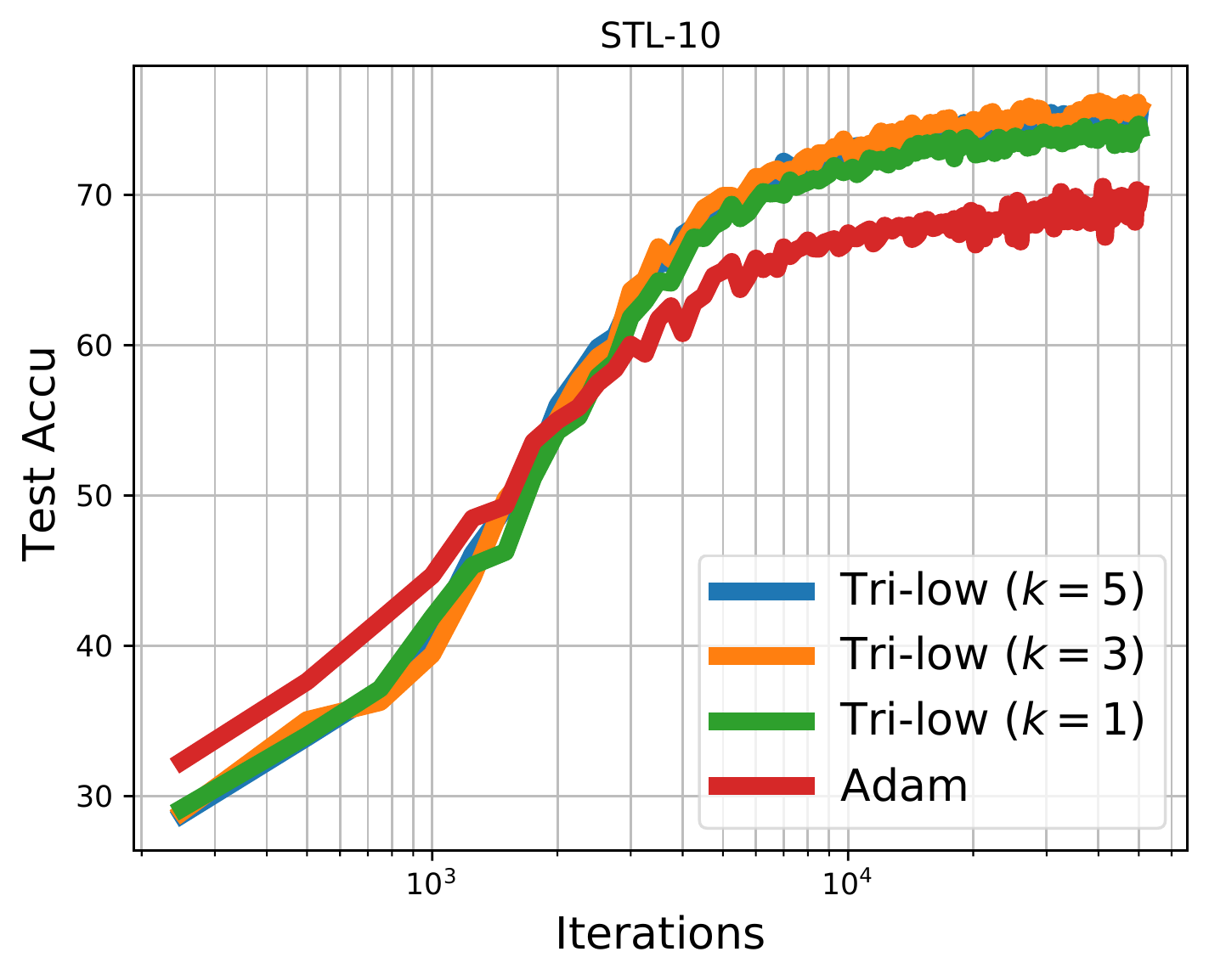}
                \caption{}
	      \label{fig:a1}
        \end{subfigure}       
	\hspace*{-0.2cm}
        \begin{subfigure}[b]{0.25\textwidth}
        	\includegraphics[width=\textwidth]{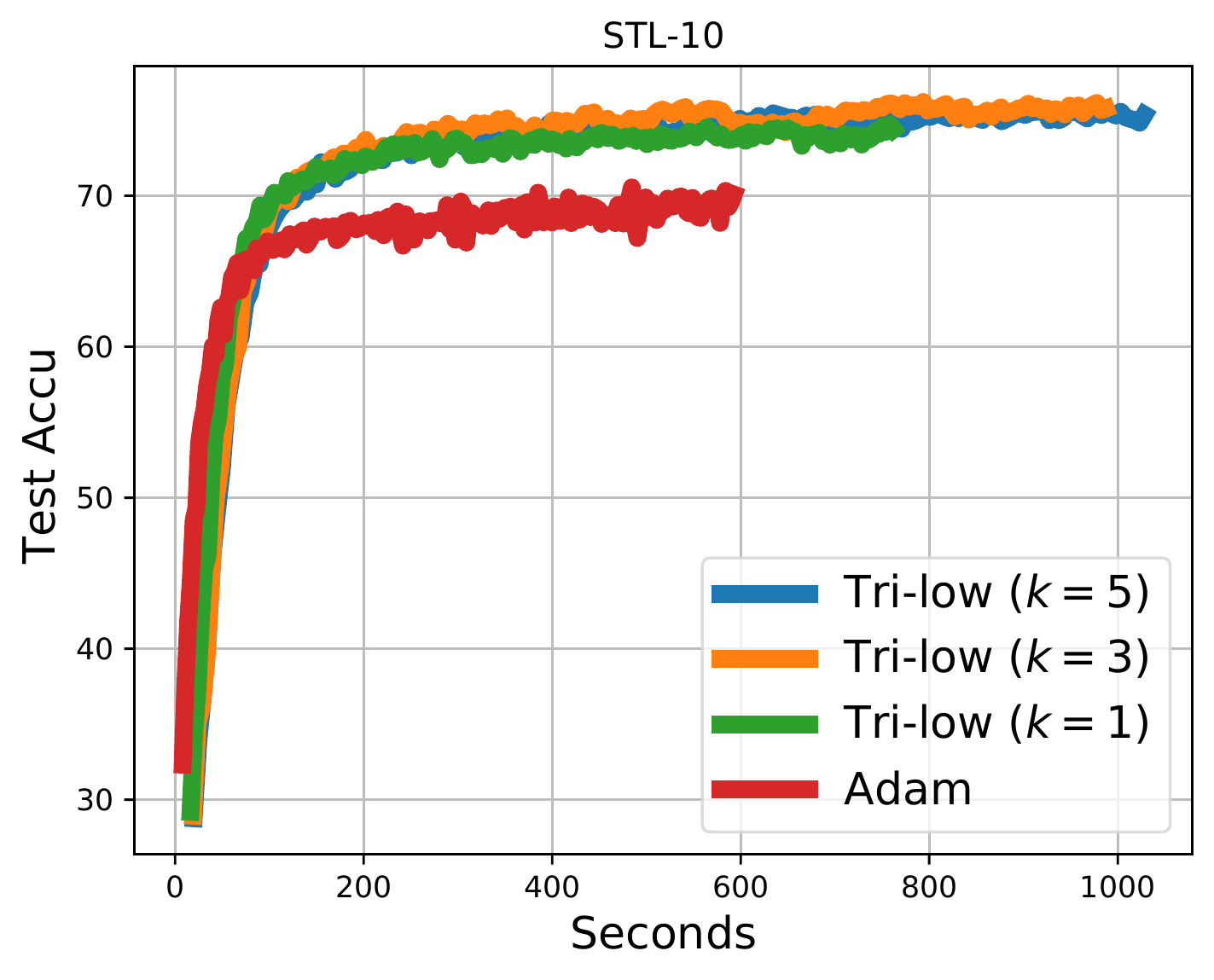}
                \caption{}
	      \label{fig:a2}
        \end{subfigure}       
     \vspace{-0.25cm}   
\hspace*{-1.2cm}
           \caption{ 
   The performances of our updates for deterministic  optimization and deep learning optimization.
Fig.~\ref{fig:b}-\ref{fig:c} show the performances using structured Newton's updates to optimize non-separable, valley-shaped, 200-dimensional functions, where our structured second-order updates only require to compute diagonal entries of Hessian and Hessian-vector products. Our updates with a lower Heisenberg structure in the precision form converge faster than BFGS and Adam.
Fig.~\ref{fig:a1}-\ref{fig:a2} shows the performances 
for optimization of a CNN using a Kronecker product group structure, where our adaptive structured updates ($O(k|\vmu|)$) have a linear iteration cost like Adam ($O(|\vmu|)$) and are automatically parallelized by Auto-Diff. Our updates achieve higher test accuracy ($75.8\%$) than Adam ($69.5\%$).
 }
\vspace{-0.2cm}
\label{fig:fig1}
\end{figure*}

\begin{figure*}[t]
	\centering
	\hspace*{-1.2cm}
	\includegraphics[width=0.25\linewidth]{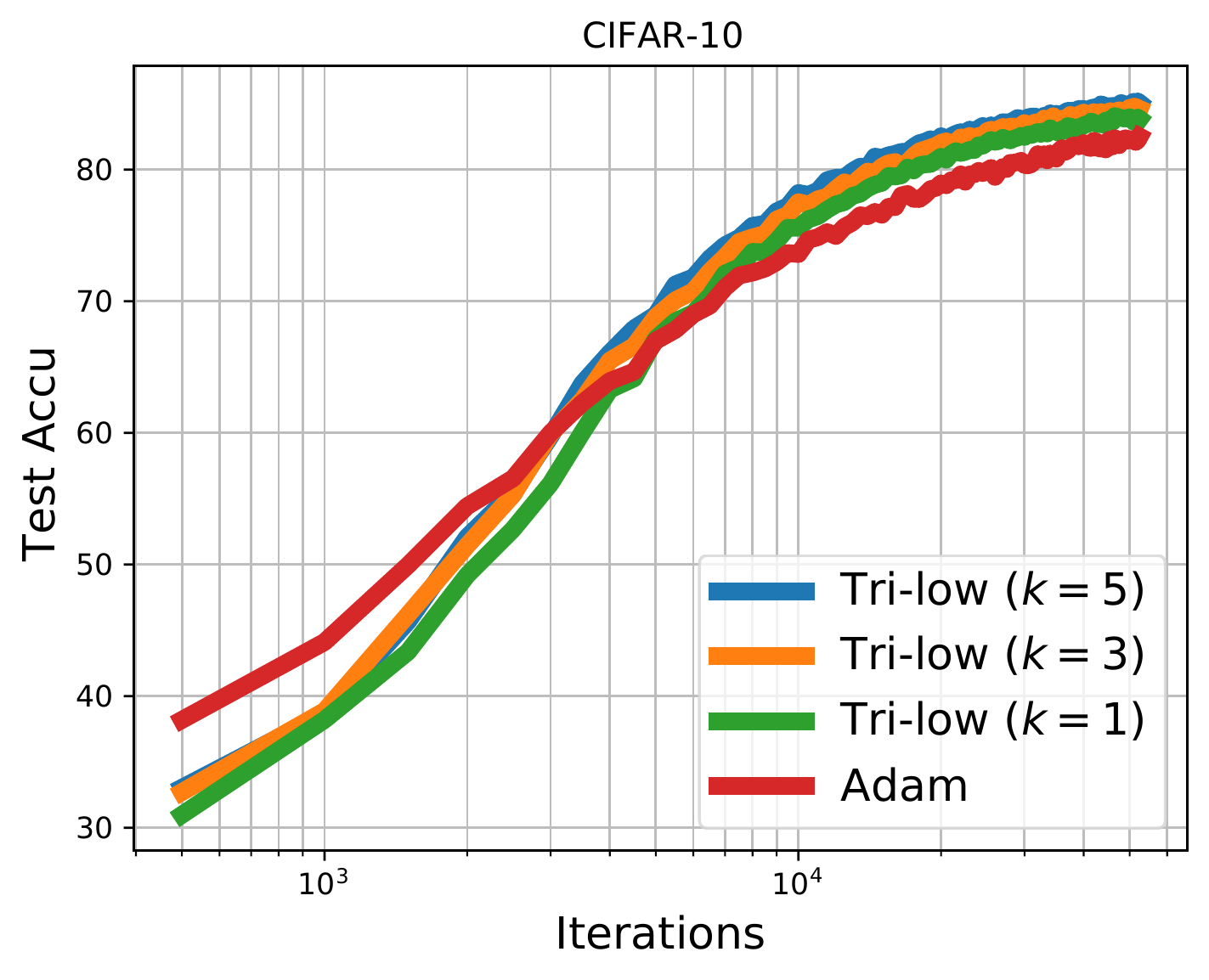}
	\includegraphics[width=0.25\linewidth]{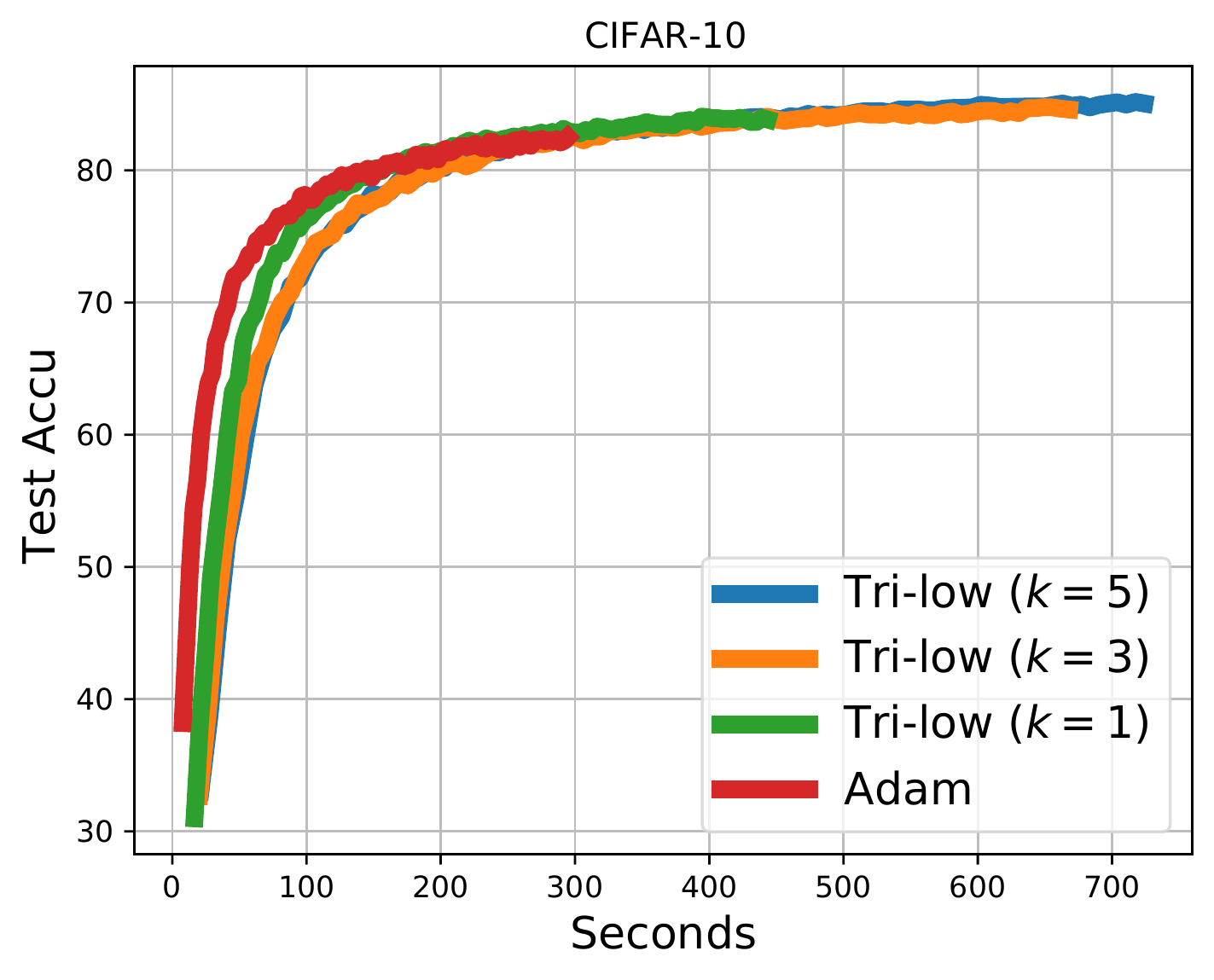}
	\includegraphics[width=0.25\linewidth]{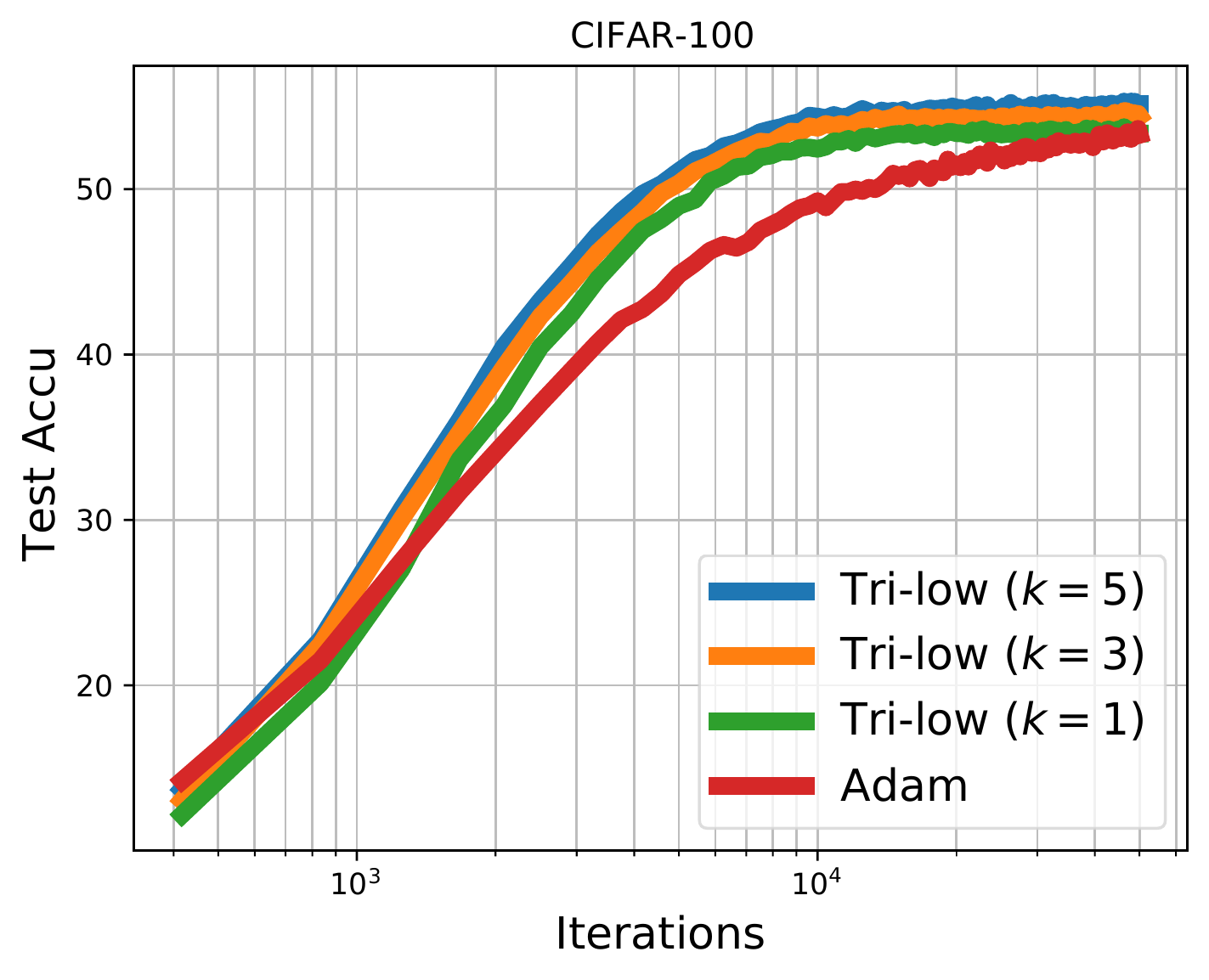}
	\includegraphics[width=0.25\linewidth]{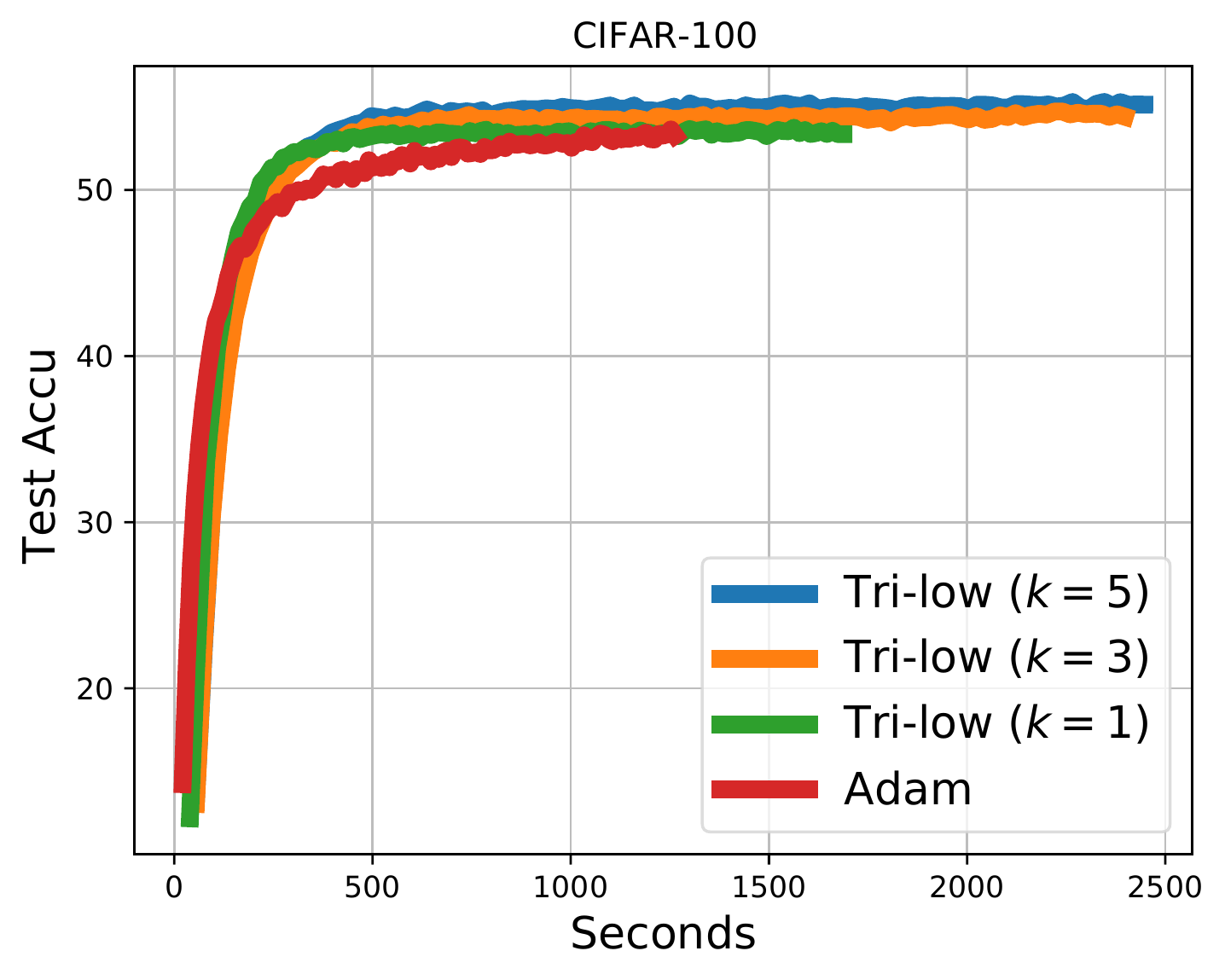}
	\hspace*{-1.2cm}
	\vspace{-0.45cm}
	\caption{
	   The performances of our updates on the same CNN model on two more datasets.  Our updates achieve higher test accuracy 
	   ($85.0\%$ on ``CIFAR-10'', $55.2\%$ on ``CIFAR-100'') than Adam ($82.3\%$ on ``CIFAR-10'', $53.3\%$ on ``CIFAR-100'').
	   }
	\label{fig:dnn}
\vspace{-0.2cm}
\end{figure*}

\subsection{Structured covariances}
\label{sec:tri_group}
To construct structured covariances, we will use structured restrictions of 
 $\mathrm{GL}^{p\times p}$. Note that $\mathrm{GL}^{p\times p}$ is a \emph{general linear group} (GL group) \citep{Group-notes}.
 Structured restrictions give us subgroups of $\mathrm{GL}^{p\times p}$.
We will first discuss a block triangular group.
More structures are illustrated in Fig.~\ref{fig:groups}.

${\cal{B}_\text{up}}(k)$ denotes
the  block upper-triangular $p$-by-$p$ matrice space, where $k$ is the block size with $0 \leq  k \leq p$,  $d_0=p-k$, and ${\cal D}^{d_0 \times d_0}_{++}$ is a diagonal and invertible matrice space.
\begin{align*}
{\cal{B}_{\text{up}}}(k)  = \Big\{ 
\begin{bmatrix}
\vB_A &  \vB_B  \\
 \mathbf{0} & \vB_D
      \end{bmatrix} \Big| & \vB_A \in \mathrm{GL}^{k \times k},\,\,
 \vB_D  \in{\cal D}^{d_0 \times d_0}_{++}  \Big\}
\end{align*} 
${\cal{B}_{\text{up}}}(k)$ is indeed a  \emph{matrix (Lie) group} closed under matrix multiplication. 
Its members are all invertible matrices.

By the structured NGD method,
we define another set for ${\cal{B}_{\text{up}}}(k)$,
where ${\cal D}^{d_0 \times d_0}$ is the diagonal matrix space.
\begin{align*}
{\cal{M}_{\text{up}}}(k)  = \Big\{ 
\begin{bmatrix}
\vM_A &  \vM_B  \\
 \mathbf{0} & \vM_D
      \end{bmatrix} \Big| &  \vM_A \in{\cal S}^{k \times k}, \,\,
 \vM_D  \in{\cal D}^{d_0 \times d_0} \Big\}
\end{align*}

\vspace{-0.1cm}
Set ${\cal{M}_{\text{up}}}(k)$, which is indeed a Lie sub-algebra, is designed for  ${\cal{B}_{\text{up}}}(k)$
so that $\vh(\vM) \in {\cal{B}_{\text{up}}}(k)$ whenever $\vM \in {\cal{M}_{\text{up}}}(k)$.

Now, consider a $p$-dimensional Gaussian distribution $q(\vlat)=\gauss(\vlat|\vmu,\vS^{-1})$, where $\vS \in \mathcal{S}_{++}^{p\times p}$ is the precision.
We consider the following covariance structure $\vS=\vB\vB^T$, where $\vB \in {\cal{B}_{\text{up}}}(k)$ is a member of the group.
Using the structured  NGD method,  the update  for \eqref{eq:problem} is  expressed as:
\begin{equation}
\begin{split}
\vmu_{t+1} & \leftarrow \vmu_{t} - \beta \vS_t^{-1} \vg_{\mu_t}  \\
\vB_{t+1} & \leftarrow   \vB_t \vh \rnd{ \beta \vC_{\text{up}} \odot \kappa_{\text{up}}\big( 2 \vB_t^{-1} \vg_{\Sigma_t} \vB_t^{-T} \big) }
\end{split}
\label{eq:eq_group_ngd}
\end{equation}

\vspace{-0.4cm}
where $\odot$ is the element-wise product, $\kappa_{\text{up}}(\vX)$ extracts non-zero entries of   ${\cal{M}_{\text{up}}}(k)$ from $\vX$ so that $\kappa_{\text{up}}(\vX) \in {\cal{M}_{\text{up}}}(k)$ and $\vC_{\text{up}}$ is a constant matrix defined below. 
\begin{align*}
 \vC_{\text{up}} = 
 \begin{bmatrix}
\half \vJ_A & \vJ_B   \\
 \mathbf{0} & \half \vI_D
      \end{bmatrix}  \in {\cal{M}_{\text{up}}}(k)
\end{align*}

\vspace{-0.4cm}
where 
 $\vJ $ is a matrix of ones and 
factor $\half$ appears in the symmetric part of ${\vC}_{\text{up}}$. 

Update \eqref{eq:eq_group_ngd} preserves the group structure: $\vB_{t+1} \in {\cal{B}_{\text{up}}}(k)$ if $\vB_k \in {\cal{B}_{\text{up}}}(k)$.
When $k=p$, ${\cal{B}_{\text{up}}}(k)  $ becomes the invertible matrix space $\mathrm{GL}^{p\times p}$. 
Therefore,
the update in \eqref{eq:eq_group_ngd} recovers the update in \eqref{eq:ngd_aux_param} with a complete covariance structure when $k=p$.
Similarly, the update in \eqref{eq:eq_group_ngd} becomes a NGD update with a diagonal covariance  when $k=0$.
When $0<k<p$, \eqref{eq:eq_group_ngd} becomes a  NGD update with a structured covariance.

We obtain an  update for problem \eqref{eq:opt} by  approximating the expectations in \eqref{eq:eq_group_ngd} at  $\vmu$ and using  \eqref{eq:gauss_stein_2nd}, where $\vS=\vB\vB^T$.
\begin{tcolorbox}[enhanced,colback=white,%
    colframe=red!75!black, attach boxed title to top right={yshift=-\tcboxedtitleheight/2, xshift=-1.25cm}, title=A structured second-order update for \eqref{eq:opt}, coltitle=red!75!black, boxed title style={size=small,colback=white,opacityback=1, opacityframe=0}, size=title, enlarge top initially by=-\tcboxedtitleheight/2]
    \vspace{0.2cm}
\begin{equation}
    \begin{split}
   \mspace{-102.5mu}\vmu_{t+1} &  \mspace{-4.5mu} \leftarrow \mspace{-4.5mu} \vmu_t - \beta \vS_{t}^{-1} \nabla_\mu \ell( \vmu_t) \\
   \mspace{-102.5mu}\vB_{t+1} & \mspace{-4.5mu} \leftarrow \mspace{-4.5mu}
     \vB_t \vh \big( \beta \vC_{\text{up}}  \mspace{-4.5mu}\odot \kappa_{\text{up}}\big(  \vB_t^{-1} \nabla_\mu^2 \ell( \vmu_t)  \vB_t^{-T}  \mspace{-6.5mu}- \mspace{-6.5mu}\gamma \vI \big) \big)
    \mspace{-20.5mu}
    \end{split}
    \mspace{-20.5mu}
 \label{eq:stru_newton}
    \mspace{-20.5mu}
 \end{equation} 
\end{tcolorbox}

We can similarly show that
this update  \eqref{eq:stru_newton}
is invariant to any transformation $\vK^T \in {\cal{B}_{\text{up}}}(k)$. Therefore, the update enjoys a \emph{group structural invariance} since
${\cal{B}_{\text{up}}}(k)$ is a group.

Thanks to the sparsity of group ${\cal{B}_{\text{up}}}(k)$  \citep{lin2021tractable}, the update in \eqref{eq:stru_newton} has low time complexity $O(k^2 p)$. 
We compute
$\vS^{-1} \nabla_\mu \ell( \vmu)$  and  $\vB \vh(\vM)$ in $O(k^2 p)$ since  $\vB$ and $\vh(\vM)$ are in ${\cal{B}_{\text{up}}}(k)$.
We compute/approximate diagonal entries of Hessian $\nabla_\mu^2 \ell( \vmu)$ and use $k$ Hessian-vector-products for non-zero entries of $\kappa_{\text{up}}\big(  \vB^{-1} \nabla_\mu^2 \ell( \vmu)  \vB^{-T}\big)$ (see Eq. (54) in Appx J.1.7 of \citet{lin2021tractable}).
We store non-zero entries of $\mathbf{B}$ with space complexity $O((k+1)p)$.

\citet{lin2021tractable} show that
there is a block arrowhead structure~\citep{o1990computing} over $\vS_{\text{up}}=\vB\vB^T$: 

\vspace{-0.35cm}
\begin{align*}
\vS_{\text{up}}
&=  \begin{bmatrix}
\vB_A  \vB_A^T + \vB_B \vB_B^T & \vB_B \vB_D \\
\vB_D \vB_B^T & \vB_D^2
      \end{bmatrix}
\end{align*}

\vspace{-0.35cm}
and over $\vSigma_{\text{up}}=\vS_{\text{up}}^{-1}$, which has a low-rank structure:

\vspace{-0.35cm}
\begin{align*}
\vSigma_{\text{up}} &= \vU_k \vU_k^T +  
      \begin{bmatrix}
      \mathbf{0}  & \mathbf{0} \\
 \mathbf{0}      &  \vB_D^{-2}
      \end{bmatrix};\,\,\,\, 
      \vU_k =  \begin{bmatrix}
-\vB_A^{-T} \\
\vB_D^{-1} \vB_B^T \vB_A^{-T}
      \end{bmatrix} 
\end{align*}

\vspace{-0.35cm}
where $\vU_k$ is a rank-$k$ matrix since $\vB_A^{-T}$ is invertible.

%

Similarly, we  define a lower-triangular group ${\cal{B}_{\text{low}}}(k)$.

\vspace{-0.35cm}
\begin{align*}
{\cal{B}_{\text{low}}}(k)  = \Big\{ 
\begin{bmatrix}
\vB_A & \mathbf{0}   \\
 \vB_C & \vB_D
      \end{bmatrix} \Big| & \vB_A \in \mathrm{GL}^{k \times k},\,\,
 \vB_D  \in{\cal D}^{d_0 \times d_0}_{++}  \Big\}
\end{align*} 

\vspace{-0.35cm}
The update obtained from the structured NGD method with a structure $\vB \in {\cal{B}_{\text{low}}}(k)$ has a low-rank structure in precision $\vS_{\text{low}}=\vB\vB^T$.
Likewise, the update  with a structure  $\vB \in {\cal{B}_{\text{up}}}(k)$ has a low-rank structure in covariance $\vS_{\text{up}}^{-1}=(\vB\vB^T )^{-1}$.
We can show that they are structured second-order updates for \eqref{eq:opt} with a structural invariance when we approximate the expectations at the mean. 

A more flexible structure is to construct a \emph{hierarchical structure} inspired by the Heisenberg group~\cite{HeisenbergdDim-2018} by replacing a diagonal group\footnote{${\cal D}^{d_0 \times d_0}_{++}$ is indeed a diagonal matrix group.} in $\vB_D$ with a block triangular group, where $0\leq k_1+k_2 \leq p$ and $d_0=p-k_1-k_2$

\vspace{-0.35cm}
\begin{align*}
 {\cal B}_{\text{up}} (k_1,k_2)
  = \Big\{  \begin{bmatrix}
\vB_A & \vB_B \\
\mathbf{0} & \vB_D
      \end{bmatrix} \Big| 
      \vB_D = \begin{bmatrix}
      \vB_{D_1} & \vB_{D_2}\\
      \mathbf{0} & \vB_{D_4}
     \end{bmatrix}
 \Big\}
\end{align*}

 \vspace{-0.35cm}
where 
$\vB_A \in \mathrm{GL}^{k_1 \times k_1}$,
$\vB_{D_1}  \in{\cal D}_{++}^{d_0 \times d_0}$,
$\vB_{D_4} \in \mathrm{GL}^{k_2 \times k_2}$.


If the Hessian $\nabla_\mu^2 \ell(\vmu)$ has a model-specific structure, 
we could use a customized group to capture such a structure in the precision.
For example, the Hessian of layer-wise matrix weights of a NN admits a Kronecker form.
We can use a \emph{Kronecker product group} to capture such structure.
This group  can reduce the time complexity from the quadratic complexity to a linear one  in $k$ (see Appx.~I.1 of \citet{lin2021tractable} and Fig.~\ref{fig:fig1}).
By approximating the expectations  at the mean and the Hessian by the Gauss-Newton, the structured NGD method  gives us a {\color{red} \emph{structured adaptive update}} for NNs.
This  update also enjoys a group structural invariance.
Existing structured adaptive methods such as KFAC~\citep{martens2015optimizing} and  Shampoo~\citep{gupta18shampoo} do not have the invariance.
Moreover,
our update
is different from existing NG methods for NN.
These existing NG methods use the empirical Fisher defined by $\ell$ while ours uses the
 the exact Fisher of the Gaussian $q$.
 Our method does not have the issue
  studied by \citet{kunstner2019limitations}.

We could obtain  more structured second-order updates by
using many subgroups (e.g., invertible circulant matrix groups, invertible triangular-Toeplitz matrix groups)
of the  $\mathrm{GL}^{p \times p}$ group and groups obtained from existing groups via the \emph{group conjugation} by an element of the orthogonal group.

\vspace{-0.2cm}
\section{Numerical Results}
\label{sec:results}


\vspace{-0.1cm}
\subsection{Structured Second-order Optimization}
We consider 200-dimensional ($p=200$), non-separable,  valley-shaped test functions for optimization:
Rosenbrock: 
$ \ell_{\text{rb}}(\vlat) =\frac{1}{p} \sum_{i=1}^{p-1}\big[100(\lat_{i+1}-\lat_i)^2 + (\lat_i-1)^2 \big]$,
and Dixon-Price:
$ \ell_{\text{dp}}(\vlat) =\frac{1}{p}\big[ (\lat_i-1)^2 + \sum_{i=2}^{p} i(2\lat_i^2 - \lat_{i-1})^2 \big]$.
We test our structured second-order updates for \eqref{eq:opt}, where we  set $\gamma =1$ in our updates.
We consider these  structures in the precision $\vS$:
 the upper triangular structure (denoted by ``Tri-up''),
 the lower triangular structure (denoted by ``Tri-low''),
 the upper Heisenberg structure (denoted by ``Hs-up''),
and the lower Heisenberg structure (denoted by ``Hs-low''), where 
second-order information is used.
For our updates, we compute Hessian-vector products and diagonal entries of the Hessian without
 computing the whole Hessian.
We consider baseline methods: the BFGS method from SciPy and the Adam optimizer, where the step-size is tuned for Adam.
Figure \ref{fig:b}-\ref{fig:c} show the performances of all methods\footnote{Empirically, we find out that a {\emph{lower structure}} in the precision  performs better than an upper structure for {optimization} tasks including optimization for neural networks.
For variational inference, the trend is opposite.  
},
where our updates with a lower Heisenberg structure converge faster than BFGS and Adam.

\vspace{-0.1cm}
\subsection{Optimization for Deep Learning}
\label{sec:opt_dl}
We consider a CNN model with 9 hidden layers. 
For a smooth objective, we use average pooling and GELU~\citep{hendrycks2016gaussian} as activation functions.
We employ $L_2$ regularization with weight $10^{-2}$.
We  set $\gamma =1$ in our updates.
We train the model with our structured adaptive updates (see Appx.~I of \citet{lin2021tractable}) for matrix weights at each layer of NN
on
datasets ``STL-10'', ``CIFAR-10'', ``CIFAR-100''.
We use a Kronecker product group structure of two lower-triangular groups (referred to as ``Tri-low'') for computational complexity reduction.
We train the model with mini-batches.
We compare our updates to Adam, where the step-size for each method is tuned by grid search.
We use the same initialization and hyper-parameters in all methods.
We report results in terms of test accuracy, where we average the results over 5 runs with distinct random seeds.
From Figures~\ref{fig:fig1}-\ref{fig:dnn}, we can see our structured updates have a linear iteration cost like Adam while achieve higher test accuracy.

\vspace{-0.2cm}
\section{Conclusion}
We propose
structured second-order updates for unconstrained optimization and
structured adaptive updates for NNs. 
These updates seem to be promising. 
An interesting direction is to evaluate these updates in  large-scale settings.

\section*{Acknowledgements}
WL is supported by a UBC International Doctoral Fellowship.
This research was partially supported by the Canada CIFAR AI Chair Program.

\bibliography{refs}
\bibliographystyle{icml2021}

\end{document}